\documentclass[10pt,harvard]{article}

\usepackage{acl2002}
\usepackage{xspace}
\usepackage{amssymb}

\title{\vspace{-75pt}
{\normalsize {\it \hfill Appears in  Proc. 2002
Conf. on Empirical Methods in Natural Language Processing (EMNLP)}} \\ \mbox{}\\Thumbs up?  \Sentiment Classification using Machine Learning Techniques}

\author{Bo Pang  \and Lillian Lee \\ Department of Computer Science \\ Cornell University
\\ Ithaca, NY 14853 USA \\ {\tt \{pabo,llee\}@cs.cornell.edu}
\And
Shivakumar Vaithyanathan \\
IBM Almaden Research Center\\
650 Harry Rd. \\
San Jose, CA  95120 USA\\
{\tt shiv@almaden.ibm.com} 
}

\newcommand{\omt}[1]{}
\newcommand{\set}[1]{\{#1\}}

\newcounter{alttablelinecount}  
\newcommand{\alttableline}{\stepcounter{alttablelinecount} (\thealttablelinecount)}
\newcommand{\assign}{:=}
\newcommand{\best}[1]{{\bf #1}}
\newcommand{\class}{c}
\newcommand{\condprob}[2]{P(#1~|~#2)}
\newcommand{\featurecount}{n}
\newcommand{\ffn}[4]{F_{#1,#2}(#3,#4)}  
\newcommand{\ffnname}[2]{F_{#1,#2}} 
\newcommand{\meabbrev}{{\rm ME}}
\newcommand{\mecondprob}[2]{P_{\meabbrev}(#1~|~#2)}
\newcommand{\mecondprobname}{P_{\meabbrev}}
\newcommand{\meshort}{MaxEnt\xspace}
\newcommand{\nbcondprob}[2]{P_{{\rm NB}}(#1~|~#2)}

\newcommand{\proc}{Proc. of the\xspace}
\newcommand{\sample}[1]{``{\sf #1}''} 
\newcommand{\sentiment}{sentiment\xspace}
\newcommand{\sentiments}{sentiments\xspace}
\newcommand{\Sentiment}{Sentiment\xspace}
\newcommand{\svmlight}{$SVM^{light}$\xspace}
\newcommand{\word}[1]{``#1''} 

\begin{document}
\maketitle

\begin{abstract}

We consider the problem of classifying documents not by topic, but by
overall sentiment, e.g., determining whether a review is positive or
negative. Using movie reviews as data, we find that standard machine
learning techniques definitively outperform human-produced baselines.
However, the three machine learning methods we employed (Naive
Bayes, maximum entropy classification, and support vector machines) do
not perform as well on sentiment classification as on traditional topic-based
categorization. We conclude by examining factors that make the
sentiment classification problem more challenging.

\end{abstract}

\section{Introduction}

Today, very large amounts of information are available in on-line
documents.  As part of the effort to better organize this information
for users, researchers have been actively investigating the problem of
automatic text categorization.  

The bulk of such work has focused on {\em topical} categorization,
attempting to sort documents according to their subject matter (e.g.,
sports vs.~politics). However, recent years have seen rapid growth in
on-line discussion groups and review sites (e.g., the {New York
Times}' Books web page) where a crucial characteristic of the posted
articles is their {\em sentiment}, or overall opinion towards the
subject matter --- for example, whether a product review is positive or
negative.  Labeling these articles with their \sentiment would provide
succinct summaries to readers; indeed, these labels are part of the
appeal and value-add of such sites as {www.rottentomatoes.com}, which
both labels movie reviews that do not contain explicit rating
indicators and normalizes the different rating schemes that individual
reviewers use.  \Sentiment classification would also be helpful in
business intelligence applications (e.g. MindfulEye's Lexant
system\footnote{http://www.mindfuleye.com/about/lexant.htm}) and
recommender systems (e.g., \newcite{Terveen+al:97a},
\newcite{Tatemura:00a}), where user input and feedback could be
quickly summarized; indeed, in general, free-form survey responses
given in natural language format could be processed using \sentiment
categorization.  Moreover, there are also potential applications to
message filtering; for example, one might be able to use \sentiment
information to recognize and discard ``flames''\cite{Spertus:97a}.

In this paper, we examine the effectiveness of applying machine
learning techniques to the \sentiment classification problem.  A
challenging aspect of this problem that seems to distinguish it from
traditional topic-based classification is that while topics are often
identifiable by keywords alone, \sentiment can be expressed in a more
subtle manner.  For example, the sentence \word{How could anyone sit
through this movie?} contains no single word that is obviously negative.
(See Section \ref{sec:discuss} for more examples).  Thus, \sentiment
seems to require more {\em understanding} than the usual topic-based
classification.  So, apart from presenting our results obtained via
machine learning techniques, we also analyze the problem to gain a
better understanding of how difficult it is.

\section{Previous Work}

This section briefly surveys previous work
on {non}-topic-based text categorization.

One area of research concentrates on classifying documents
according to their {\em source} or {\em source style}, with statistically-detected
stylistic variation \cite{Biber:88a} serving as an important cue.  Examples include
author, publisher (e.g., the {\em New York Times} vs. {\em The Daily News}), 
native-language background, and ``brow'' 
(e.g., high-brow vs. ``popular'', or low-brow)
\cite{Mosteller+Wallace:84a,Argamon+Koppel+Avneri:98b,Tomokiyo+Jones:01a,Kessler+Nunberg+Schuetze:97a}.

Another, more related area of research is that of determining the {\em
genre} of texts; subjective genres, such as ``editorial'', are often
one of the possible categories
\cite{Karlgren+Cutting:94a,Kessler+Nunberg+Schuetze:97a,Finn+Kushmerick+Smyth:02a}.
Other work
explicitly attempts to find features indicating that subjective
language is being used  \cite{Hatzivassiloglou+Wiebe:00a,Wiebe+Wilson+Bell:01a}.
But, while techniques for genre categorization and subjectivity
detection can help us {\em recognize} documents that
express an opinion, they do not address our specific {\em
classification} task of determining what that opinion actually is.

Most previous research on sentiment-based classification has been at
least partially knowledge-based.  Some of this work focuses on
classifying the semantic orientation of individual words
or phrases,
 using
linguistic heuristics or a pre-selected set of seed words
\cite{Hatzivassiloglou+McKeown:97a,Turney+Littman:02a}.  Past work on
sentiment-based categorization of entire documents has often involved either
the use of models inspired by  cognitive linguistics \cite{Hearst:92a,Sack:94a} or the
manual or semi-manual construction of discriminant-word lexicons
\cite{Huettner+Subasic:00a,Das+Chen:01a,Tong:01a}.
Interestingly,
our baseline experiments, described in Section \ref{sec:closer}, show
that humans may not always have the best intuition for choosing
discriminating words.

Turney's \shortcite{Turney:02a} work on classification of reviews is
perhaps the closest to ours.\footnote{Indeed, although our choice of
title was completely independent of his, our selections were eerily
similar.} He applied a specific unsupervised learning technique
based on 
the mutual information between document
phrases and the words ``excellent'' and ``poor'',  where the mutual
information is computed using statistics
gathered by a search engine.  In contrast, we utilize several completely
prior-knowledge-free supervised machine learning methods, with the
goal of understanding the inherent difficulty of the task.

\section{The Movie-Review Domain}
\label{sec:domain}

For our experiments, we chose to work with movie reviews.  This domain
is experimentally convenient because there are large on-line
collections of such reviews, and because reviewers often summarize
their overall \sentiment with a machine-extractable {\em rating}
indicator, such as a number of stars; hence, we did not need to
hand-label the data for supervised learning or evaluation purposes.
We also note that \newcite{Turney:02a} found movie reviews
to be the most difficult of several domains for sentiment classification,
reporting an accuracy of 65.83\% on a 120-document set (random-choice
performance: 50\%).
But we stress that the machine learning methods and features we use
are not specific to movie reviews, and should be easily applicable to
other domains as long as sufficient training data exists.

Our data source was the Internet Movie Database (IMDb) archive of
the {\tt rec.arts.movies.reviews} newsgroup.\footnote{http://reviews.imdb.com/Reviews/}~We selected
only reviews where the author rating was expressed either with stars
or some numerical value (other conventions varied too widely to
allow for automatic processing).
Ratings were automatically extracted and converted into one of three
categories: positive, negative, or neutral. For the work described in
this paper, we concentrated only on discriminating between positive
and negative \sentiment.
To avoid domination 
of the corpus
by a small number of prolific
reviewers, we imposed a limit of fewer than 20 reviews per author per
\sentiment category, yielding a corpus of 752 negative and 1301
positive reviews, with a total of 144 reviewers represented.  This
dataset will be available on-line at {\footnotesize {\tt
http://www.cs.cornell.edu/people/pabo/movie-review-data/}}.

\begin{figure*}[ht]
\begin{center}
{\small
\begin{tabular}{|c|l||l|l|}	\hline
	 & 	Proposed word lists	
&Accuracy &Ties\\ 	\hline \hline
Human 1	& positive: {\it dazzling, brilliant, phenomenal, excellent,
	fantastic}			& 58\% & 75\%	\\ 	
	& negative: {\it suck, terrible, awful, unwatchable, hideous}			&&	\\	\hline 
Human 2		&positive: {\it gripping, mesmerizing, riveting,
spectacular, cool},	& 64\%	& 39\%	\\ 
& ~~~~~~~{\it awesome, thrilling, badass, excellent, moving, exciting}
			&	& \\	
	&negative: {\it bad, cliched, sucks, boring, stupid, slow}				&&\\	\hline
\end{tabular}
}
\end{center}

\caption{\label{fig:baseline1} Baseline results for human
word lists.  Data: 700 positive and 700
negative reviews.}
\end{figure*}

\section{A Closer Look At the Problem}
\label{sec:closer}

\begin{figure*}[ht]
\begin{center}
{\small
\begin{tabular}{|c|l||c|c|}	\hline
	& Proposed word lists	 & Accuracy & Ties\\ 	\hline \hline
Human 3~+~stats	& positive: {\em love, wonderful, best, great, superb, still, beautiful}				& 69\% &16\%	\\
	& negative: {\em bad, worst, stupid, waste, boring, ?, !} 	&&\\	\hline
\end{tabular}
}
\end{center}

\caption{\label{fig:baseline2} Results for baseline using
introspection and simple statistics of the data (including {\em test} data).}
\end{figure*}

Intuitions seem to differ as to the difficulty of the sentiment
detection problem.  An expert on 
using machine learning for text
categorization predicted relatively low performance for automatic
methods.  On the other hand, it seems that
distinguishing positive from negative reviews is relatively easy for
humans, especially in comparison to the standard text categorization
problem, where topics can be closely related.  One might also suspect
that there are certain words people tend to use to express strong
sentiments, so that it might suffice to simply produce a list of such
words by introspection and rely on them alone to classify the texts.

To test this latter hypothesis, we asked two graduate
students in computer science to (independently) choose 
good indicator words for positive and negative
\sentiments in movie reviews.  Their selections, shown in
Figure \ref{fig:baseline1}, seem intuitively plausible.  We then
converted their responses into simple decision procedures that
essentially count the number of the proposed positive and negative
words in a given document.  We applied these procedures to
uniformly-distributed data, so that the random-choice baseline result
would be 50\%.
As shown in Figure \ref{fig:baseline1}, the accuracy 
--- percentage of documents classified correctly ---
for the
human-based classifiers were 58\% and 64\%, respectively.\footnote{Later experiments using these words as
features for machine learning methods did not yield better
results.}
Note that the tie rates --- percentage of documents where the two 
sentiments were rated equally likely --- are quite high\footnote{This
is largely due to 0-0 ties.}
 (we chose a tie breaking 
policy that maximized the accuracy of the baselines). 

While the tie rates suggest that the brevity of the human-produced
lists is a factor in the relatively poor performance results, it is
not the case that size alone necessarily limits accuracy.  Based on a
very preliminary examination of frequency counts in the entire corpus (including {\em test} data)
plus introspection, we created a list of seven positive and seven negative
words (including punctuation), shown in Figure \ref{fig:baseline2}. As
that figure indicates, using these words raised the accuracy to 69\%.
Also, although this third list is of comparable length to the other
two, it has a much lower tie rate of 16\%.  We further observe that
some of the items in this third list, such as \word{?} or \word{still},
would probably not have been proposed as possible candidates merely
through introspection, although upon reflection one sees their
merit (the question mark tends to occur in sentences like
\word{What was the director thinking?}; \word{still} appears in sentences
like \word{Still, though, it was worth seeing}).

We conclude from these preliminary experiments that it is worthwhile
to explore corpus-based techniques, rather than relying on prior
intuitions, to select good indicator features and to perform \sentiment
classification in general.  These experiments also provide us with
baselines for experimental comparison; in particular, the third
baseline of 69\% might actually be considered somewhat difficult to
beat, since it was achieved by examination of the test
data (although our examination was rather cursory; we do not
claim that our list was the optimal set of fourteen words).

\section{Machine Learning Methods}

Our aim in this work was to examine whether it suffices to treat \sentiment
classification simply as a special case of topic-based categorization (with
the two ``topics'' being positive \sentiment and negative \sentiment),
or whether special \sentiment-categorization methods need to be developed.
We experimented with three standard algorithms: Naive Bayes
classification, maximum entropy classification, and support vector
machines.  The philosophies behind these three algorithms are quite
different, but each has been shown to be effective in previous text
categorization studies.

To implement these machine learning algorithms on our document data,
we used the following standard bag-of-features framework.  Let
$\set{f_1, \ldots, f_m}$ be a predefined set of $m$ {\em features}
that can appear in a document; examples include the word
\word{still} or the bigram \word{really stinks}.  Let
$\featurecount_i(d)$ be the number of times $f_i$ occurs in document
$d$.  Then, each document $d$ is represented by the document vector
$\vec{d} \assign
(\featurecount_1(d), \featurecount_2(d), \ldots, \featurecount_m(d))$.

\subsection{Naive Bayes}

One approach to text classification is to assign to a given document
$d$ the class $\class^* = \arg\max_c \condprob{\class}{d}$.  We derive
the {\em Naive Bayes} (NB) classifier by first observing that by
Bayes' rule,
$$\condprob{\class}{d} = \frac{P(\class)
\condprob{d}{\class}}{P(d)},$$ where $P(d)$ plays no role in selecting
$\class^*$. To estimate the term $\condprob{d}{\class}$, Naive
Bayes decomposes it
by assuming the $f_i$'s are conditionally independent given $d$'s
class:
$$\nbcondprob{\class}{d} \assign \frac{P(\class) \left(\prod_{i=1}^m \condprob{f_i}{\class} ^ {\featurecount_i(d)}\right)}{P(d)}.$$
Our training method consists of relative-frequency estimation of $P(\class)$ and
$\condprob{f_i}{\class}$, using add-one smoothing.

Despite its simplicity and the fact that its conditional independence
assumption clearly does not hold in real-world situations, Naive
Bayes-based text categorization still tends to perform surprisingly
well \cite{Lewis:98a}; indeed, \newcite{Domingos+Pazzani:97a} show
that Naive Bayes is optimal for certain problem classes with highly
dependent features.  On the other hand, more sophisticated algorithms
might (and often do) yield better results; we examine two such
algorithms next.

\subsection{Maximum Entropy}

Maximum entropy classification (\meshort, or \meabbrev, for short)  is
an alternative technique which has proven effective in a number of
natural language processing applications
\cite{Berger+DellaPietra+DellaPietra:96a}.
\newcite{Nigam+Lafferty+McCallum:99a} show that it sometimes, but not
always, outperforms Naive Bayes at standard text classification.  Its
estimate of $\condprob{\class}{d}$ takes the following exponential
form:
$$\mecondprob{\class}{d} \assign \frac {1}{Z(d)} \exp\left(\sum_{i}
\lambda_{i,\class} \ffn{i}{\class}{d}{\class}\right)\;,$$ where
$Z(d)$ is a normalization function.
$\ffnname{i}{\class}$ is a {\em feature/class  function} for feature
$f_i$ and class $\class$, defined as follows:\footnote{We use a restricted
definition of feature/class functions so that \meshort
relies on the same sort of feature information as Naive Bayes.}
$$\ffn{i}{\class}{d}{\class'} 
\assign \cases{1, & $\featurecount_i(d) > 0$ {and $\class'=\class$} \cr
	 0 & otherwise\cr}.
$$
For instance, a particular feature/class function might fire if and only if
the bigram \word{still hate} appears and the document's \sentiment is
hypothesized to be negative.\footnote{The 
dependence on class is
necessary for parameter induction. See
\newcite{Nigam+Lafferty+McCallum:99a} for additional motivation.}
Importantly, unlike Naive Bayes, \meshort
makes no assumptions about the relationships between features, and so
might potentially perform better when conditional independence
assumptions are not met.

The $\lambda_{i,\class}$'s are feature-weight parameters; inspection
of the definition of $\mecondprobname$ shows that a large
$\lambda_{i,\class}$ means that $f_i$ is considered a strong indicator
for class $\class$.
The parameter values are set so as to maximize the entropy of the
induced distribution (hence the classifier's name) subject to the
constraint that the expected values of the feature/class functions
with respect to the model are equal to their expected values with
respect to the training data: the underlying philosophy is that we
should choose the model making the fewest assumptions about the data
while still remaining consistent with it, which makes intuitive sense.
We use ten iterations of the improved iterative scaling algorithm
\cite{DellaPietra+DellaPietra+Lafferty:97a} for parameter training
(this was 
a
sufficient 
number of iterations
for convergence of training-data accuracy),
together with a Gaussian prior to prevent overfitting
\cite{Chen+Rosenfeld:00a}.

\subsection{Support Vector Machines}

Support vector machines (SVMs) have been shown to be highly effective
at traditional text categorization, generally outperforming Naive
Bayes \cite{Joachims:98a}.  They are {\em large-margin}, rather than
probabilistic, classifiers, in contrast to Naive Bayes and \meshort.
In the two-category case, the basic idea behind the training procedure
is to find a hyperplane, represented by vector $\vec{w}$, that not
only separates the document vectors in one class from those in the
other, but for which the separation, or {\em margin}, is as large as
possible.
This search corresponds to a constrained
optimization problem;
letting $\class_j \in \set{1,-1}$ (corresponding to positive and negative)
be the correct class of document $d_j$, the solution can be written as
$$\vec{w} := \sum_j \alpha_j \class_j \vec{d_j}, \;\; \alpha_j \ge 0,$$
where the $\alpha_j$'s are obtained by solving a dual optimization
problem.  Those $\vec{d_j}$ such that $\alpha_j$ 
is greater than zero
are called {\em
support vectors}, since they are the only document vectors
contributing to $\vec{w}$.  Classification of test instances consists
simply of determining which side of $\vec{w}$'s hyperplane they fall
on.

We used Joachim's \shortcite{Joachims:99a} \svmlight
package\footnote{http://svmlight.joachims.org} for training and
testing, with all parameters set to their default values, after first
length-normalizing the document vectors, as is standard (neglecting to normalize generally hurt performance slightly).

\begin{figure*}[th]
\begin{center}
\begin{tabular}{|c|c|c|c||c|r|r|r|}\hline
	      &	Features	&\# of		&frequency or  	&~NB~ 	&~ME~		&~SVM ~\\ 
	      &			&features	& presence?	&   	&            	& 	\\ \hline\hline
\alttableline & unigrams	& 16165		&freq. 	&\best{78.7} & N/A & 72.8 	\\ \hline\hline
\alttableline & unigrams	& ''		&pres.	&81.0 & 80.4 & \best{82.9}	 \\ \hline \hline
\alttableline & unigrams+bigrams& 32330		&pres.	&80.6 & 80.8 & \best{82.7}		\\ \hline
\alttableline & bigrams 	& 16165		&pres.	&77.3 & \best{77.4} & 77.1		\\ \hline\hline
\alttableline & unigrams+POS 	& 16695		&pres.	&81.5 & 80.4 & \best{81.9} \\ \hline
\alttableline & adjectives	&2633		&pres.	&77.0 & \best{77.7} & 75.1		\\ \hline
\alttableline &top 2633 unigrams&2633		&pres.	&80.3 & 81.0 & \best{81.4}	\\ \hline\hline
\alttableline &unigrams+position 	& 22430	&pres.	&81.0 & 80.1 & \best{81.6}\\ \hline 
\end{tabular}
\end{center}
\caption{\label{fig:results-plain} Average three-fold cross-validation accuracies, in percent. Boldface:
best performance for a given setting (row).  Recall that our baseline
results ranged from 50\% to 69\%.}
\end{figure*}

\section{Evaluation}

\subsection{Experimental Set-up}

We used documents from the movie-review corpus described in Section
\ref{sec:domain}.  To create a data set with uniform class
distribution (studying the effect of skewed class distributions was
out of the scope of this study), we randomly selected 700
positive-sentiment and 700 negative-sentiment documents.  We then
divided this data into three equal-sized folds, maintaining balanced
class distributions in each fold. (We did not use a larger number of
folds due to the slowness of the \meshort training procedure.) All
results reported below, as well as the baseline results from
Section \ref{sec:closer}, are the average three-fold cross-validation
results on this data (of course, the baseline algorithms had no
parameters to tune).

To prepare the documents, we automatically removed the rating indicators and
extracted the textual information from the original HTML document
format, treating punctuation as separate lexical items.  No stemming
or stoplists were used.

One unconventional step we took was to attempt to model the
potentially important contextual effect of negation: clearly
\word{good} and \word{not very good} indicate opposite \sentiment
orientations.
Adapting a technique of \newcite{Das+Chen:01a}, we added the tag {\tt
NOT\_} to every word between a negation word (\word{not},
\word{isn't}, \word{didn't}, etc.) and the first punctuation mark
following the negation word. (Preliminary experiments indicate that
removing the negation tag had a negligible, but on average slightly
harmful, effect on performance.)

For this study, we focused on features based on unigrams (with
negation tagging) and bigrams.  Because training \meshort is expensive
in the number of features, we limited consideration to (1) the 16165
unigrams appearing at least four times in our 1400-document
corpus (lower count cutoffs did
not yield significantly different results), and (2) the 16165 bigrams
occurring most often in the same data (the selected bigrams all
occurred at least seven times).  Note that we did not add negation
tags to the bigrams, since we consider bigrams (and $n$-grams in
general) to be an orthogonal way to incorporate context.

\subsection{Results}

\paragraph{Initial unigram results} 

The classification accuracies resulting from using only unigrams as
features are shown in line (1) of Figure \ref{fig:results-plain}.  As
a whole, the machine learning algorithms clearly surpass the
random-choice baseline of 50\%.  They also handily beat our two
human-selected-unigram baselines of 58\% and 64\%, and, furthermore,
perform well in comparison to the 69\% baseline achieved via limited
access to the test-data statistics, although the improvement
in the case of SVMs is not so large.

On the other hand, in topic-based classification, all three classifiers
have been reported to use bag-of-unigram features to achieve accuracies of
90\% and above for particular categories \cite{Joachims:98a,Nigam+Lafferty+McCallum:99a}\footnote{\newcite{Joachims:98a}
used stemming and stoplists; in some of their experiments,
\newcite{Nigam+Lafferty+McCallum:99a}, like us, did not.}
--- and such results are for settings with more than two classes.
This provides suggestive evidence that \sentiment categorization is
more difficult than topic classification, which corresponds to the
intuitions of the text categorization expert mentioned
above.\footnote{We
could not perform the natural experiment of attempting topic-based
categorization on our data 
because the only obvious topics would be
the film being reviewed;  unfortunately, in our
data, the maximum number of reviews per movie is 27, too small for meaningful results.}  Nonetheless, we still
wanted to investigate ways to improve our \sentiment categorization
results; these experiments are reported below.

\paragraph{Feature frequency vs. presence} 

Recall that we represent each document $d$ by a feature-count vector
($\featurecount_1(d), \ldots, \featurecount_m(d))$.  However, the
definition of the \meshort feature/class functions $\ffnname{i}{c}$
only reflects the presence or absence of a feature, rather than
directly incorporating feature frequency. In order to
investigate whether reliance on frequency information could account
for the higher accuracies of Naive Bayes and SVMs, we binarized the
document vectors, setting $\featurecount_i(d)$ to 1 if and only
feature $f_i$ appears in $d$, and reran Naive Bayes and \svmlight on
these new vectors.\footnote{Alternatively, we could have tried
integrating frequency information into \meshort.  However, feature/class
functions are traditionally defined as binary
\cite{Berger+DellaPietra+DellaPietra:96a}; hence, explicitly incorporating
frequencies would require different functions for each count (or count
bin), making training impractical.  But cf. \cite{Nigam+Lafferty+McCallum:99a}.}  

As can be seen from line (2) of Figure \ref{fig:results-plain}, better
performance ({\em much} better performance for SVMs) is achieved by
accounting only for feature presence, not feature frequency.
Interestingly, this is in direct opposition to the observations of
\newcite{McCallum+Nigam:98a} with respect to Naive Bayes topic
classification.  We speculate that this indicates a difference between
\sentiment and topic categorization --- perhaps due to topic being
conveyed mostly by particular content words that tend to be repeated --- but
this remains to be verified.  In any event, as a result of this
finding, we did not incorporate frequency information into Naive Bayes
and SVMs in any of the following experiments.

\paragraph{Bigrams} In addition to looking specifically for negation
words in the context of a word, we also studied the use of bigrams to
capture more context in general.
Note that bigrams and unigrams are surely not conditionally
independent, meaning that the feature set they comprise violates Naive Bayes'
conditional-independence assumptions; on the other
hand, recall that this does not imply that Naive Bayes will
necessarily do poorly \cite{Domingos+Pazzani:97a}.

Line (3) of the results table shows that bigram information does not
improve performance beyond that of unigram presence, although adding
in the bigrams does not seriously impact the results, even for Naive
Bayes.  This would not rule out the possibility that bigram presence
is as equally useful a feature as unigram presence; 
in fact, \newcite{Pedersen:01a} found that bigrams alone can be effective
features for word sense disambiguation. However,  comparing line (4) to
line (2) shows that  relying just on bigrams causes accuracy
to decline by as much as 5.8 percentage points. 
Hence, if context is
in fact important, as our intuitions suggest, bigrams are not
effective at capturing it in our setting.

\paragraph{Parts of speech} We also experimented with appending 
POS tags to every word via
Oliver Mason's Qtag
program.\footnote{http://www.english.bham.ac.uk/staff/oliver/\-soft\-ware/\-tagger/\-index.htm}
This serves as a crude form of word sense disambiguation
\cite{Wilks+Stevenson:98a}: for example, it would distinguish
the different usages of \word{love} in \word{I love this movie}
(indicating \sentiment orientation) versus \word{This is a love story}
(neutral with respect to \sentiment).  However, the effect of this
information seems to be a wash: as depicted in line (5) of Figure
\ref{fig:results-plain}, the accuracy improves slightly for Naive
Bayes but declines for SVMs, and the
performance of \meshort is unchanged.

Since adjectives have been a focus of previous work in \sentiment
detection \cite{Hatzivassiloglou+Wiebe:00a,Turney:02a}\footnote{Turney's
\shortcite{Turney:02a} unsupervised algorithm
uses bigrams containing an adjective or an adverb.}, we looked at the
performance of using adjectives alone.  Intuitively, we might expect
that adjectives carry a great deal of information regarding a
document's \sentiment; indeed, the human-produced lists from Section
\ref{sec:closer} contain almost no other parts of speech.  Yet, the
results, shown in line (6) of Figure \ref{fig:results-plain}, are
relatively poor:  
the 2633 adjectives provide less useful
information than unigram presence. Indeed, line (7) shows that simply
using the 2633 most frequent unigrams is a better choice, yielding
performance comparable to that of using (the presence of) all 16165
(line (2)). This may imply that applying explicit feature-selection
algorithms on unigrams could improve performance.

\paragraph{Position} An additional intuition we had was
that the position of 
a word in the text might make a difference:
movie reviews, in particular, might begin with an overall \sentiment
statement, proceed with a plot discussion, and conclude by summarizing
the author's views.
As a rough approximation to determining this kind of structure, we
tagged each word according to whether it appeared in the first quarter, 
last quarter, or middle half of the document\footnote{We tried a few
other settings, e.g., first third vs. last third vs middle third, and
found them to be less effective.}.  The results (line (8)) didn't differ greatly from using unigrams alone, but more refined notions of position might be more successful.

\section{Discussion}
\label{sec:discuss}

The results produced via machine learning techniques are quite good in
comparison to the human-generated baselines discussed in Section
\ref{sec:closer}.  In terms of relative performance, Naive Bayes tends to 
do the worst and SVMs tend to do the best, although the differences aren't very large.

On the other hand, we were not able to achieve
accuracies on the \sentiment classification problem comparable to
those reported for standard topic-based categorization, despite the
several different types of features we tried. Unigram presence
information turned out to be the most effective; in fact, none of the
alternative features we employed provided consistently better
performance once unigram presence was incorporated.  Interestingly,
though, the superiority of presence information in comparison to
frequency information in our setting contradicts previous observations
made in topic-classification work \cite{McCallum+Nigam:98a}.  

What accounts for these two  differences --- difficulty and types of
information proving useful ---  between topic and \sentiment
classification, and how might we improve the latter?  To answer these
questions, we examined the data further. (All examples below are drawn from
the full 2053-document corpus.)

As it turns out, a common phenomenon in the documents was a kind of
``thwarted expectations'' narrative, where the author sets up a
deliberate contrast to earlier discussion: for example, \sample{This
film should be brilliant. It sounds like a great plot, the actors are
first grade, and the supporting cast is good as well, and Stallone is
attempting to deliver a good performance. However, it can't hold up}
or
\sample{I hate the Spice Girls. ...[3 things the author hates about
them]...  Why I saw this movie is a really, really, really long story,
but I did, and one would think I'd despise every minute of it.
But... Okay, I'm really ashamed of it, but I enjoyed it.  I mean, I
admit it's a really awful movie ...\omt{act wacky as hell...}the ninth
floor of hell...\omt{a cheap ass movie...}The plot is such a mess that it's
terrible.  But I loved it.} 
\footnote{This phenomenon is related to another common theme, that of ``a good
actor trapped in a bad movie'':
\sample{AN AMERICAN WEREWOLF IN PARIS is a failed attempt... 
Julie Delpy is far too good for this movie. She imbues Serafine with
spirit, spunk, and humanity\omt{, which gives us an emotional stake in the
character's fate}. This isn't necessarily a good thing, since it
prevents us from relaxing and enjoying AN AMERICAN WEREWOLF IN PARIS
as a completely mindless, campy entertainment experience. Delpy's
injection of class into an otherwise classless production raises the
specter of what this film could have been with a better script and a
better cast  \omt{surrounding her.  ...Delpy's previous credits include such
memorable ventures as}... She was radiant, charismatic, and effective
....}
}

In these examples, a human would easily detect the true sentiment of
the review, but bag-of-features classifiers would presumably find
these instances difficult, 
since there are many words indicative of the 
opposite sentiment to that of the entire review. Fundamentally, it seems that some form of
discourse analysis is necessary (using more sophisticated techniques
than our positional feature mentioned above), or at least some way of
determining the focus of each sentence, so that one can decide when
the author is talking about the film itself. (\newcite{Turney:02a}
makes a similar point, noting that for reviews, ``the whole is not necessarily the
sum of the parts''.)  Furthermore, it seems
likely that this thwarted-expectations rhetorical device will appear
in many types of texts (e.g., editorials) devoted to expressing an
overall opinion about some topic.  Hence, we believe that an important
next step is the identification of features indicating whether
sentences are on-topic (which is a kind of co-reference problem); we
look forward to addressing this challenge in future work.

\subsection*{Acknowledgments}
We thank Joshua Goodman, Thorsten Joachims, Jon Kleinberg, Vikas
Krishna, John
Lafferty, Jussi Myllymaki, Phoebe Sengers, Richard
Tong, Peter Turney, and the anonymous reviewers for many valuable
comments and helpful suggestions, and Hubie Chen and Tony Faradjian
for participating in our baseline experiments. Portions of this work
were done while the first author was visiting IBM Almaden.  This paper is based
upon work supported in part by the National Science Foundation under
ITR/IM grant IIS-0081334. Any opinions, findings, and conclusions or
recommendations expressed above are those of the authors and do not
necessarily reflect the views of the National Science Foundation.

\bibliographystyle{agsm}

\begin{thebibliography}{}

\bibitem[\protect\citename{Argamon-Engelson \bgroup et al.\egroup }1998]{Argamon+Koppel+Avneri:98b}
Shlomo Argamon-Engelson, Moshe Koppel, and Galit Avneri.
\newblock 1998.
\newblock Style-based text categorization: What newspaper am {I} reading?
\newblock In {\em \proc AAAI Workshop on Text Categorization}, pages 1--4.

\bibitem[\protect\citename{Berger \bgroup et al.\egroup
  }1996]{Berger+DellaPietra+DellaPietra:96a}
Adam~L. Berger, Stephen~A. {Della Pietra}, and Vincent~J. {Della Pietra}.
\newblock 1996.
\newblock A maximum entropy approach to natural language processing.
\newblock {\em Computational Linguistics}, 22(1):39--71.

\bibitem[\protect\citename{Biber}1988]{Biber:88a}
Douglas Biber.
\newblock 1988.
\newblock {\em Variation across Speech and Writing}.
\newblock Cambridge University Press.

\bibitem[\protect\citename{Chen and Rosenfeld}2000]{Chen+Rosenfeld:00a}
Stanley Chen and Ronald Rosenfeld.
\newblock 2000.
\newblock A survey of smoothing techniques for {ME} models.
\newblock {\em IEEE Trans. Speech and Audio Processing}, 8(1):37--50.

\bibitem[\protect\citename{Das and Chen}2001]{Das+Chen:01a}
Sanjiv Das and Mike Chen.
\newblock 2001.
\newblock Yahoo! for {Amazon}: Extracting market sentiment from stock message
  boards.
\newblock In {\em \proc 8th Asia Pacific Finance Association Annual Conference
  (APFA 2001)}.

\bibitem[\protect\citename{{Della Pietra} \bgroup et al.\egroup
  }1997]{DellaPietra+DellaPietra+Lafferty:97a}
Stephen {Della Pietra}, Vincent {Della Pietra}, and John Lafferty.
\newblock 1997.
\newblock Inducing features of random fields.
\newblock {\em IEEE Transactions on Pattern Analysis and Machine Intelligence},
  19(4):380--393.

\bibitem[\protect\citename{Domingos and Pazzani}1997]{Domingos+Pazzani:97a}
Pedro Domingos and Michael~J. Pazzani.
\newblock 1997.
\newblock On the optimality of the simple {B}ayesian classifier under zero-one
  loss.
\newblock {\em Machine Learning}, 29(2-3):103--130.

\bibitem[\protect\citename{Finn \bgroup et al.\egroup
  }2002]{Finn+Kushmerick+Smyth:02a}
Aidan Finn, Nicholas Kushmerick, and Barry Smyth.
\newblock 2002.
\newblock Genre classification and domain transfer for information filtering.
\newblock In {\em \proc European Colloquium on Information Retrieval Research},
  pages 353--362, Glasgow.

\bibitem[\protect\citename{Hatzivassiloglou and
  McKeown}1997]{Hatzivassiloglou+McKeown:97a}
Vasileios Hatzivassiloglou and Kathleen McKeown.
\newblock 1997.
\newblock Predicting the semantic orientation of adjectives.
\newblock In {\em \proc 35th ACL/8th EACL}, pages 174--181.

\bibitem[\protect\citename{Hatzivassiloglou and
  Wiebe}2000]{Hatzivassiloglou+Wiebe:00a}
Vasileios Hatzivassiloglou and Janyce Wiebe.
\newblock 2000.
\newblock Effects of adjective orientation and gradability on sentence
  subjectivity.
\newblock In {\em Proc. of COLING}.

\bibitem[\protect\citename{Hearst}1992]{Hearst:92a}
Marti Hearst.
\newblock 1992.
\newblock Direction-based text interpretation as an information access
  refinement.
\newblock In Paul Jacobs, editor, {\em Text-Based Intelligent Systems}.
  Lawrence Erlbaum Associates.

\bibitem[\protect\citename{Huettner and Subasic}2000]{Huettner+Subasic:00a}
Alison Huettner and Pero Subasic.
\newblock 2000.
\newblock Fuzzy typing for document management.
\newblock In {\em {ACL} 2000 Companion Volume: Tutorial Abstracts and
  Demonstration Notes}, pages 26--27.

\bibitem[\protect\citename{Joachims}1998]{Joachims:98a}
Thorsten Joachims.
\newblock 1998.
\newblock Text categorization with support vector machines: Learning with many
  relevant features.
\newblock In {\em \proc European Conference on Machine Learning (ECML)}, pages
  137--142.

\bibitem[\protect\citename{Joachims}1999]{Joachims:99a}
Thorsten Joachims.
\newblock 1999.
\newblock Making large-scale {SVM} learning practical.
\newblock In Bernhard Sch\"{o}lkopf and Alexander Smola, editors, {\em Advances
  in Kernel Methods - Support Vector Learning}, pages 44--56. MIT Press.

\bibitem[\protect\citename{Karlgren and Cutting}1994]{Karlgren+Cutting:94a}
Jussi Karlgren and Douglass Cutting.
\newblock 1994.
\newblock Recognizing text genres with simple metrics using discriminant
  analysis.
\newblock In {\em Proc. of {COLING}}.

\bibitem[\protect\citename{Kessler \bgroup et al.\egroup
  }1997]{Kessler+Nunberg+Schuetze:97a}
Brett Kessler, Geoffrey Nunberg, and Hinrich Sch{\"u}tze.
\newblock 1997.
\newblock Automatic detection of text genre.
\newblock In {\em \proc 35th ACL/8th EACL}, pages 32--38.

\bibitem[\protect\citename{Lewis}1998]{Lewis:98a}
David~D. Lewis.
\newblock 1998.
\newblock Naive ({B}ayes) at forty: The independence assumption in information
  retrieval.
\newblock In {\em \proc European Conference on Machine Learning (ECML)}, pages
  4--15.
\newblock Invited talk.

\bibitem[\protect\citename{McCallum and Nigam}1998]{McCallum+Nigam:98a}
Andrew McCallum and Kamal Nigam.
\newblock 1998.
\newblock A comparison of event models for {Naive Bayes} text classification.
\newblock In {\em \proc AAAI-98 Workshop on Learning for Text Categorization},
  pages 41--48.

\bibitem[\protect\citename{Mosteller and Wallace}1984]{Mosteller+Wallace:84a}
Frederick Mosteller and David~L. Wallace.
\newblock 1984.
\newblock {\em Applied Bayesian and Classical Inference: The Case of the
  Federalist Papers}.
\newblock Springer-Verlag.

\bibitem[\protect\citename{Nigam \bgroup et al.\egroup
  }1999]{Nigam+Lafferty+McCallum:99a}
Kamal Nigam, John Lafferty, and Andrew McCallum.
\newblock 1999.
\newblock Using maximum entropy for text classification.
\newblock In {\em \proc IJCAI-99 Workshop on Machine Learning for Information
  Filtering}, pages 61--67.

\bibitem[\protect\citename{Pedersen}2001]{Pedersen:01a}
Ted Pedersen.
\newblock 2001.
\newblock A decision tree of bigrams is an accurate predictor of word sense.
\newblock In {\em \proc Second NAACL}, pages 79--86.

\bibitem[\protect\citename{Sack}1994]{Sack:94a}
Warren Sack.
\newblock 1994.
\newblock On the computation of point of view.
\newblock In {\em \proc Twelfth AAAI}, page 1488.
\newblock Student abstract.

\bibitem[\protect\citename{Spertus}1997]{Spertus:97a}
Ellen Spertus.
\newblock 1997.
\newblock Smokey: Automatic recognition of hostile messages.
\newblock In {\em Proc. of Innovative Applications of Artificial Intelligence
  (IAAI)}, pages 1058--1065.

\bibitem[\protect\citename{Tatemura}2000]{Tatemura:00a}
Junichi Tatemura.
\newblock 2000.
\newblock Virtual reviewers for collaborative exploration of movie reviews.
\newblock In {\em \proc 5th International Conference on Intelligent User
  Interfaces}, pages 272--275.

\bibitem[\protect\citename{Terveen \bgroup et al.\egroup }1997]{Terveen+al:97a}
Loren Terveen, Will Hill, Brian Amento, David McDonald, and Josh Creter.
\newblock 1997.
\newblock {PHOAKS}: {A} system for sharing recommendations.
\newblock {\em Communications of the ACM}, 40(3):59--62.

\bibitem[\protect\citename{Tomokiyo and Jones}2001]{Tomokiyo+Jones:01a}
Laura~Mayfield Tomokiyo and Rosie Jones.
\newblock 2001.
\newblock You're not from round here, are you? {Naive Bayes} detection of
  non-native utterance text.
\newblock In {\em \proc Second NAACL}, pages 239--246.

\bibitem[\protect\citename{Tong}2001]{Tong:01a}
Richard~M. Tong.
\newblock 2001.
\newblock An operational system for detecting and tracking opinions in on-line
  discussion.
\newblock Workshop note, SIGIR 2001 Workshop on Operational Text
  Classification.

\bibitem[\protect\citename{Turney and Littman}2002]{Turney+Littman:02a}
Peter~D. Turney and Michael~L. Littman.
\newblock 2002.
\newblock Unsupervised learning of semantic orientation from a
  hundred-billion-word corpus.
\newblock Technical Report EGB-1094, National Research Council Canada.

\bibitem[\protect\citename{Turney}2002]{Turney:02a}
Peter Turney.
\newblock 2002.
\newblock Thumbs up or thumbs down? {Semantic} orientation applied to
  unsupervised classification of reviews.
\newblock In {\em \proc ACL}.

\bibitem[\protect\citename{Wiebe \bgroup et al.\egroup
  }2001]{Wiebe+Wilson+Bell:01a}
Janyce~M. Wiebe, Theresa Wilson, and Matthew Bell.
\newblock 2001.
\newblock Identifying collocations for recognizing opinions.
\newblock In {\em \proc ACL/EACL Workshop on Collocation}.

\bibitem[\protect\citename{Wilks and Stevenson}1998]{Wilks+Stevenson:98a}
Yorick Wilks and Mark Stevenson.
\newblock 1998.
\newblock The grammar of sense: Using part-of-speech tags as a first step in
  semantic disambiguation.
\newblock {\em Journal of Natural Language Engineering}, 4(2):135--144.

\end{thebibliography}

\end{document}